\documentclass[10pt,twocolumn,letterpaper]{article}

\usepackage{cvpr}
\usepackage{times}
\usepackage{epsfig}
\usepackage{graphicx}
\usepackage{amsmath}
\usepackage{amssymb}
\usepackage{cite}
\usepackage{caption} 
\usepackage{subcaption} 
\usepackage[ruled,vlined,linesnumbered]{algorithm2e} 
\usepackage{flexisym} 
\usepackage{amsmath} 
\usepackage{amssymb}
\usepackage{amsmath}
\usepackage[section]{placeins}
\usepackage{comment}
\usepackage{amsmath}
\usepackage{float}
\usepackage{filecontents}
\usepackage[breaklinks=true,bookmarks=false]{hyperref}

\cvprfinalcopy 

\def\cvprPaperID{****} 

\begin{document}

\title{TinyCNN: A Tiny Modular CNN Accelerator for Embedded FPGA }
\author{Ali Jahanshahi\\
ajaha004@ucr.edu \\
University of California, Riverside
}

\maketitle

\begin{abstract}
In recent years, Convolutional Neural Network (CNN) based methods have achieved great success in a large number of applications and have been among the most powerful and widely used techniques in computer vision. However, CNN-based methods are computational-intensive and resource-consuming, and thus are hard to be integrated into embedded systems such as smart phones, smart glasses, and robots. FPGA is one of the most promising platforms for accelerating CNN, but the limited  on-chip memory size limit the performance of FPGA accelerator for CNN. 

~~In this paper, we propose a framework for designing CNN accelerator on embedded FPGA for image classification. The proposed framework provides a tool for FPGA resource-aware design space exploration of CNNs and automatically generates the hardware description of the CNN to be programmed on a target FPGA. The framework consists of three main backends; software, hardware generation, and simulation/precision adjustment. The \textit{software backend} serves as an API to the designer to design the CNN and train it according to the hardware resources that are available. Using the CNN model, \textit{hardware backend} generates the necessary hardware components and integrates them to generate the hardware description of the CNN. Finaly,  \textit{Simulation/precision adjustment} backend adjusts the inter-layer precision units to minimize the classification error. 

~~We used 16-bit fixed-point data in a CNN accelerator (FPGA) and compared it to the exactly similar software version running on an ARM processor (32-bit floating point data). We encounter about $3\%$ accuracy loss in classification of the accelerated (FPGA) version. In return, we got up to $15.75\times$ speedup by classifying with the accelerated version on the FPGA.

\end{abstract}

\section{Introduction}

The exponential growth of big data during the last decade motivates for innovative methods to extract high semantic information from raw sensor data such as videos, images and speech sequences. Among the proposed methods, Convolutional Neural Networks (CNNs) have become the de-facto standard by delivering near-human accuracy in many applications related to machine vision. While CNNs  have  been  known  to  researchers  for  decades,  they were popularized after demonstrating high accuracy at the 2012  ImageNet  recognition  challenge~\cite{krizhevsky2012imagenet}. Subsequently, CNNs have become the state-of-the-art for image classification, detection, and localization tasks. Research in CNNs and other areas of deep learning continues at a rapid pace, with hundreds of new papers published each year introducing new models and techniques.

The CNN's high performance (at classification, detection, and localization) comes at the price of a large computational cost as they require tens of GOP/s to classify a single frame. Thus, one challenge to the widespread deployment of CNNs is their  significant  demands  for  computation  and  storage  capacity. Therefore, dedicated hardware is required to accelerate their execution. Graphics Processing Units (GPUs), are the most widely used platform to implement CNNs as they offer the best performance in terms of pure computational throughput, reaching up to 11 TFLOP/s. Nevertheless, in terms of power consumption, Field-Programmable Gate Array (FPGA) solutions are known to be more energy efficient (vs GPUs). Recent  work  by Microsoft has  even  explored  cost-effective  acceleration  of deep learning on FPGAs at datacenter scale~\cite{ovtcharov2015accelerating, jahanshahi2013blokus}. There are also  efforts  in  the  academic  community  on  FPGA-based CNN  accelerators~\cite{zhang2015optimizing, qiu2016going}  as  well  as  tools  for  generating them  automatically~\cite{suda2016throughput, wang2016deepburning}.


We observe two trends which may help overcome implementing CNN on FPGAs. The first is a series of recent papers in the machine learning  community  regarding  very-low-precision CNNs. Networks with binary weights~\cite{courbariaux2015binaryconnect}, or binary weights and activations~\cite{courbariaux2016binarized, rastegari2016xnor} have in certain cases demonstrated accuracy comparable to full precision nets. Such binarized neural net-works (BNNs) may be the key to efficient deep learning on FPGA. Binarization reduces storage and memory bandwidth requirements, and replace FP operations with binary operations which can be very efficiently performed on the LUT-based FPGA fabric. Concerning the cost and effort of FPGA implementation,we see a steady improvement in FPGA design automation tools over the past decade. High-level synthesis (HLS) tools such as Xilinx Vivado HLS~\cite{cong2011high} and LegUp~\cite{canis2013legup} enable a user to write code in a high-level programming language, then algorithmically compile that code down to a register-transfer level (RTL) design specification. More recent tools such as Intel FPGA SDK for OpenCL~\cite{czajkowski2012opencl} and Xilinx SDSoC~\cite{kathail2016sdsoc} offer further automation features for generating the hardware-software interface and on-chip memory network. In the context of deep learning, these tools have the potential to critically reduce time-to-market on new accelerator designs and thus reduce the aforementioned innovation gap.

In this project, we aim to design and implement a CNN accelerator on an embedded FPGA for image classification. We use~\cite{zhao-bnn-fpga2017} as our reference that presents the design of a BNN accelerator for FPGAs. In contrast to our reference, in this work, the aim is to provide designers a \textit{general} framework to enable them design their CNN \textit{easily} and use it as an accelerator on an FPGA. Thus, we do not use BNN-based CNN in this project.

In addition to generality of the framework, we also target embedded systems that have hardware resource limitation. Compressing the CNN model is a good choice to address the resource limitations of the hardware. A straight forward way to compress a network is to reduce the bit-width for computing. This utilizes the flexibility of FPGA or ASIC design compared with GPU or CPU. It is proved to be an effective way to use 16-bits fixed-point operations in ~\cite{cavigelli2016origami} with small loss in accuracy. In our framework, we use 16-bits operation and we show that the accuracy loss is very small. 

This work proposes a general framework for design and implementation of tiny modular CNN accelerators on embedded FPGAs. Our framework automatically generates the hardware code for the designed CNN and trained with the provided software backend. The framework consists of three main components that correspond to our contributions. They are as follows:
\begin{itemize}
    \item \textbf{Software backend:} Using \textit{Python}, we provide a tool for designing and training a CNN architecture. The model (weights) of the trained network and other CNN architecture parameters are used in the hardware framework to generate the hardware description language of the CNN, which is going to be implemented on the FPGA. This back-end also provides the information needed for \textit{Simulation/precision adjustment} backend as well as checking the hardware resources for the designed CNN model.
    \item \textbf{Hardware backend:} Using \textit{CHISEL}~\cite{bachrach2012chisel}, we perform automatic generation and integration of different hardware components needed for the CNN architecture designed in the software backend. The CNN model is passed to this back-end to be used in HDL generation of the CNN. The output of this backend is the HDL code that is ready to be synthesized and programmed on an FPGA.
    \item \textbf{Simulation/precision adjustment backend:} Using \textit{Scala} testing libraries and the data passed to this backend, inter-layer precision of the generated CNN hardware will be adjusted. The CNN output error introduced by varying integer and fractional part of each layer's data is minimized by this backend.
\end{itemize}
The rest of this paper is organized as follows; we first describe the framework overall structure, then we go in details for each component of the framework. Finally, we provide the results discussing different aspects of the generated CNN for our target FPGA.
\section{Tiny CNN}
\begin{figure*}[]
        \centering
        \includegraphics[width=0.7\linewidth]{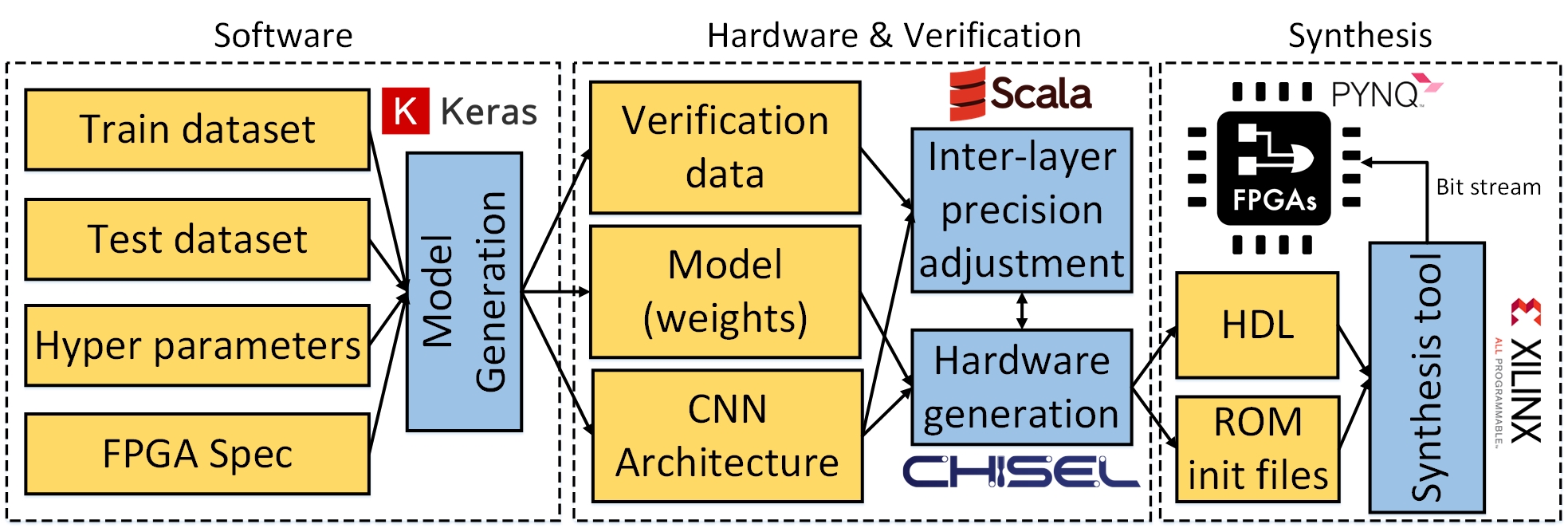}
        \caption{Tiny CNN framework.}\label{fig:framework}
\end{figure*}

CNN is a machine learning classifier that typically takes in an image and produces the probabilities of that image belonging to each output class. A typical CNN consists of a pipeline of connected \textit{layers}. Each layer takes as input a set of feature maps (\textit{fmaps}), performs some computation on them, and produces a new set of \textit{fmaps} to be fed into the next layer. The input \textit{fmaps} of the first layer is the input image. Layers may require configuration  values  known  as \textit{parameters},  which  must  first  be  determined by training the CNN offline on pre-classified data. Once the parameters are finalized, the CNN can be deployed for inference — the  classification  of  new  data  points.  For most practical machine learning applications, the first-class concerns are the accuracy and execution time of online classification. This project will thus focus on accelerating the inference task without compromising accuracy. The aim of this project is to provide a framework that automates implementation of CNNs on an embedded FPGA for image classification.  Figure~\ref{fig:framework} shows the framework we proposed for this purpose. In the following subsections, we are going to elaborate more on each component of our framework.

\subsection{Software backend}
The software backend is an API by which users are capable of designing their CNN in the software level. The user can tune their CNN with the \textit{train} and \textit{test} data sets, and finally, export their model for further processings. We used \textit{Keras} library as the underlying deep learning framework of the software backend. \textit{Keras} is a high-level neural networks API, written in Python and capable of running on top of \textit{TensorFlow}, \textit{CNTK}, or \textit{Theano}. It was developed with a focus on enabling fast experimentation. We address three challenges of designing a CNN by Software backend:

\noindent \textbf{CNN designing, training, and model generation:} In our framework, using \textit{Keras} API, we provide a template \textit{Python} class to the designers to design their CNN. Then, the designer's CNN is trained and the model is saved in order to be used by the hardware backend. Essentially, the model is the CNN weights that are going to be used as initial values of ROMs. 

\noindent \textbf{Verification data:} The software backend also inputs some random data from test dataset to CNN and captures the output of all layers of CNN with respect to that input. The collected dataset  is called verification data and is used in hardware backend for precision adjustment and verification purposes. 

\noindent \textbf{Hardware resource check:} As mentioned before, we are targeting embedded FPGA in this work. Thus, we want to squeez the whole CNN model in FPGA BRAMs. Thus, one step for designing the CNN would be checking if the CNN model (weights) fit in FPGA BRAMs. Software backend uses the target FPGA spec to check if the model fits in FPGA or not. In case the model does not fit on the target FPGA, the software backend throws an exception that shows not enough space for the model. In such cases, the designer should play around the CNN hyper parameters and make it smaller to fit on the FPGA. Fully-connected layers, due to their high number of parameters contributes to the most of the memory (BRAM) usage. 

\subsection{Hardware backend}
Hardware backend generates the hardware of the CNN which is designed in the software backend according to the inputs that are provided to this backend. We implemented all basic hardware components that are needed for CNNs. The developed hardware components are \textit{modular} and \textit{highly configurable}. They can be configured based on the CNN specifications including: data bit-width, shared or exclusive components, and number of \textit{DSP}s available for components. All of the implemented hardware components are modular that means we can simply attach them together with few lines of code just like we do in designing CNNs in software. 
\subsubsection{Convolution unit}
This unit is  the  most  critical component of the accelerator, as it will take up the vast majority of the runtime. The unit must maintain high throughput and resource efficiency. Convolution operation has multiplication and addition in its core. Since multiplication is a very expensive operation, we used the built-in \textit{DSP}s on FPGA board that are specialized for digital signal processing purposing -- mainly include multiplication and addition. 

The challenge with using the FPGA \textit{DSP}s is that there are a limited number of them on every FPGA board. The implemented Convolution unit hardware is configurable in such a way that the designer determines the number of \textit{DSP}s for this layer, and the hardware generator framework generates the convolver state machine in a way that it uses only the specified number of \textit{DSP}s. Apparently, the more \textit{DSP}s we allot to this hardware unit, the more throughput we get. The maximum number of \textit{DSP}s is the input image pixels, and the minimum is one.

Since we are targeting embedded systems, in our framework, this unit can be used by the designer of the CNN in two modes:
\begin{itemize}
    \item \textbf{Shared mode:} in the this mode, the convolution unit is shared among all layers of CNN. Sharing this units results in using less hardware resources of the FPGA, but it introduces throughput degradation. If this mode is chosen by the designer, a wrapper is generated automatically by the hardware generator backend. The wrapper acts as a resource manager for this unit by arbitrating different layers requesting to use convolution unit. 
    \item \textbf{Exclusive mode:} in the exclusive mode, for each convolution layer in the CNN a convolution unit hardware is generated, which results in higher throughput and FPGA resource usage. 
\end{itemize}{}

\subsubsection{FeedForward unit}
The input to convolution unit hardware is three lines of the input \textit{fmap} and the filter that is going to be applied on the \textit{fmap}s. Also, the output of the convolution unit hardware is one line that its size equal to the input \textit{fmap} size and it goes through activation, max pooling, and precision adjustment layer to get to the next layer. Such inter-layer data should be handled and stored in a hardware unit. 

\begin{figure}[t]
        \centering
        \includegraphics[width=0.5\linewidth]{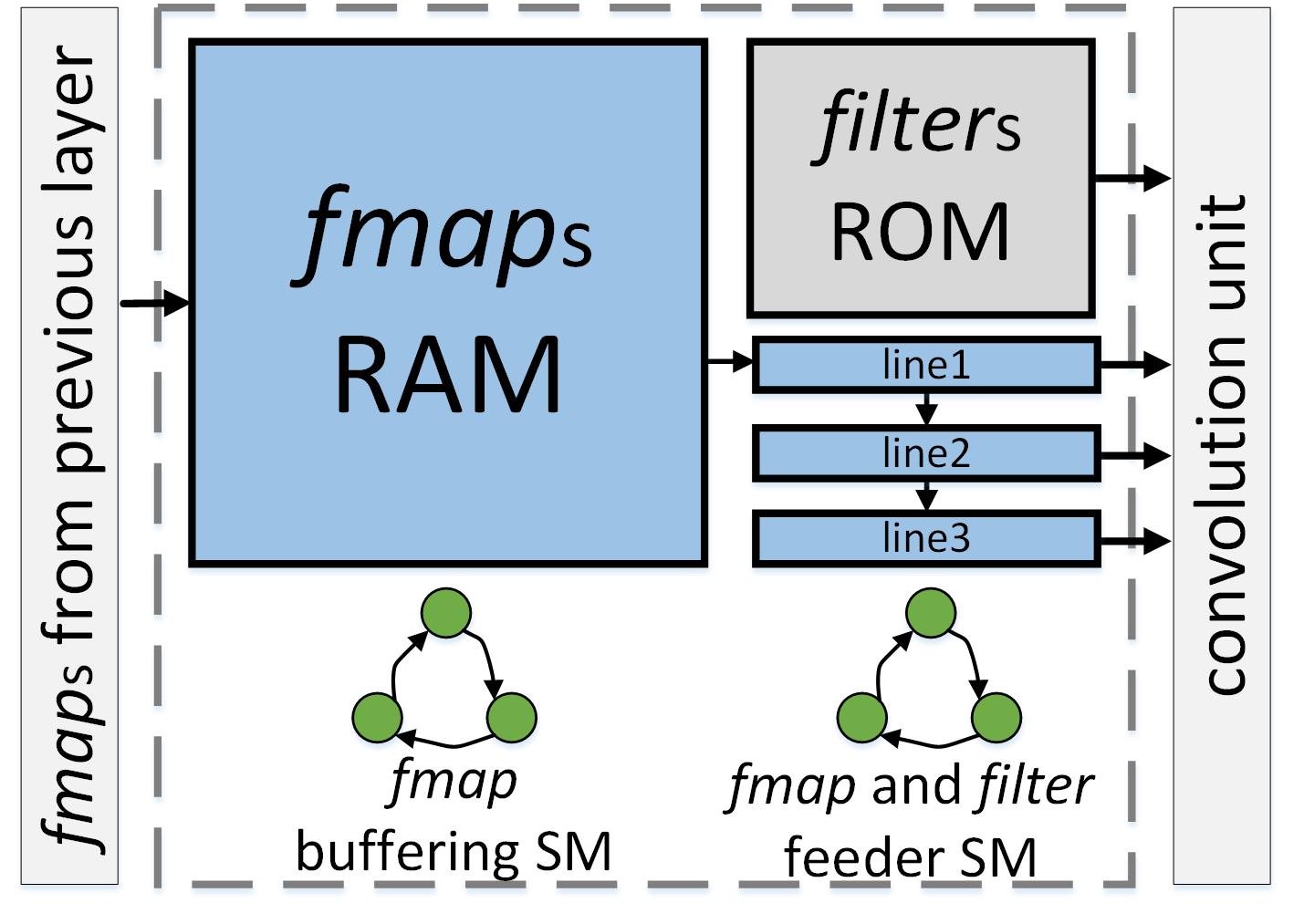}
        \caption{FeedForward unit architecture.}\label{fig:ff}
\end{figure}

FeedForward unit is responsible for handling accumulation of the input \textit{fmap}s in an inter-layer RAM by a state machine (SM), and output them to the convolution unit three line at a time in conjunction with the  \textit{filter} that should be applied to the \textit{fmap}. Figure~\ref{fig:ff} shows the structure of this hardware unit. 

As it is shown in Figure~\ref{fig:ff}, one state machine (SM) is responsible for buffering the feature maps (\textit{fmap}) that are input to this layer. This state machine is generated for each layer according to the size and number of \textit{fmap}s that are inputted to the layer. On the output side, another state machine is responsible for feeding three lines of \textit{fmaps} to the convolution unit as well as the filter that is going to applied to that. The state machine is also generated specific to the layer. Also, the RAM and ROM in Figure~\ref{fig:ff} are generated according to the \textit{fmap}s that are going to buffered in this layer and number of that layer filters. ROMs are initialized with the CNN model (weights) provided by the Software backend.

\subsubsection{Activation and Max pooling units}
In our framework, we implemented a max pooling unit and an activation hardware unit. The activation unit performs a Rectified Linear \textit{ReLU} function. The max pooling unit is configurable and is generated to perform $M \times M$ max pooling in which $M$ is specified by the designer.

\subsubsection{Inter-layer precision adjustment unit}
As we mentioned before, we are using fixed-point numbers and operations in our framework. Using fixed-point instead of floating-point introduced error to the CNN. This error propagates through the network and affects the final results adversely. Since the error propagated through the early layers has more negative effects on the output anf the range of the data propagated through the network varies from one layer to the other, we need to adjust the proportion of integer part and fractional part at the end of each layer of the network. For doing so, we inser an Inter-layer precision adjustment unit at the end of each layer. This unit is responsible for adjusting the integer and fractional part of the output data of each layer before propagating to the rest of the network. The adjustment of this unit is done by the Simulation/precision adjustment backend. 

\subsubsection{Dense or fully-connected (FC) unit}
Dense or fully-connected is also implemented as a configurable hardware unit. Since this unit mainly consists of multiplication and addition, it needs \textit{DSP} resources. In our framework, the number of DSPs and the size of the FC unit is adjustable according to the designer needs. The weights for this layer are stored in ROMs. So, they are generated based on the weights provided by Software backend and used in the synthesis phase of the generate hardware.

\subsection{Simulation and Precision Adjustment Backend}
In order to adjust the fixed-point precision of the CNN, we need to propagate several test inputs to the CNN implemented in the hardware. The test points are generated by software backend based on the user's input dataset for training the CNN. The data consists of the input to the CNN and the output of all layers that are produced for that image. The simulation backend uses \textit{Scala} language testing libraries to input the verification data to the generated hardware. Then all the data from the output of each layer of the generated CNN hardware is collected, and the difference of the collected data and the data provided by the software backend are calculated. Then, this backend adjusts the bit-width of the Inter-layer precision adjustment unit of each layer based to minimize the error generated by each layer. 

This sequence of propagating data, collecting data, calculating error, and adjusting the precision units is performed until the network merges to a minimum output error. At the end, the Simulation and precision adjustment backend adjusts all integer and fractional parts of the layers based on the verification data that are provided by the Software backend. 

\section{Experimental Results}
We evaluate our design on a \textit{PYNQ} board~\cite{pynq} that is an open-source project from \textit{Xilinx} that makes it easy to design embedded systems with \textit{Xilinx Zynq SoC}. \textit{PYNQ} uses a low-cost \textit{Xilinx Zynq-7000 SoC} containing an \textit{XC7Z020} FPGA alongside an \textit{ARM  Cortex-A9}  embedded  processor. On the software side of \textit{PYQN}, using the Python language and libraries, designers can exploit the benefits of programmable logic and microprocessors in \textit{Zynq} to build more capable and exciting embedded systems. We make  use  of  \textit{Xilinx  SDSoC  2016.4}  as  the  primary  design tool,  which  leverages  \textit{Vivado  HLS}  and  Vivado  to  perform the  actual  synthesis and programming the FPGA implementation. We used CIFAR-10 dataset to train and test the CNN. We converted the images to grey to have one-channel input due to our a small FPGA board. 

\begin{table}
\centering
\caption{The CNN architecture generated by the framework.}\label{table:1}
\begin{tabular}{lccc}
\hline
 Layer type& Output size & Output \textit{fmap}s  & Param \#  \\ 
\hline
\hline
 Conv&  (32, 32) & 32 & 320  \\
 Activation& (32, 32)&32  & 0    \\
 Max pool&   (16, 16)&32 &  0 \\
 \hline
 Conv&  (16, 16)&64 &  18496  \\
 Activation&  (16, 16)& 64& 0    \\
 Max pool&  (8, 8)&64& 0     \\
 \hline
Conv&  (8, 8)&128 & 73856    \\
 Activation&  (8, 8)&128 & 0    \\
 Max pool&  (4, 4)& 128 & 0    \\
 \hline
 Conv&  (4, 4)& 128&  147584    \\
 Activation&  (4, 4)& 128 & 0    \\
 Max pool& (2, 2)& 128& 0     \\
 \hline
 Dense & (1, 100) &  &  51300   \\
 Activation& (1, 100) &  &  0    \\
 Dense & (1, 10) &  &  1010   \\
 Activation& (1, 10) &  & 0  \\
 \hline
 \hline
 Total&  &  & 292,566  \\ 
 \hline
\end{tabular}
\end{table}

Using the software backend we designed and trained a CNN. Table~\ref{table:1} shows the architecture of the CNN we targeted for classifying CIFAR-10 dataset. We used $3\times3$ filters for convolutions. The reason behind choosing such a small CNN is that we are targeting a small FPGA as our hardware platform and any larger CNN would not pass the checking resource limitation phase of the Software backend. Also, we wanted to show the accuracy difference between the accelerated version on FPGA and the software version. 

In order to compare the software version and the version implemented on FPGA, we classified CIFAR-10 test dataset for both software model on \textit{ARM  Cortex-A9} which is the embedded  processor of our FPGA and on the TinyCNN accelerator with two modes. Table~\ref{table:2} shows the comparison averaged results per image. \textit{SW} is the \textit{Python}-based software classifier ran on ARM processor of the board (650MHz). \textit{HW-SM} is the CNN accelerator generated in shared mode (sharing one convolution unit among all layers), and \textit{HW-EM} is the CNN accelerator generated in exclusive mode (one convolution for each layer). 

We see that the effect of using 16-bit fixed-point data on the precision of the classification is negligible. We can also see that using exclusive convolution hardware units improves the runtime by avoiding stalls due to access contention for using convolution layer by different layers.

\begin{figure}[t]
        \centering
        \includegraphics[width=0.8\linewidth]{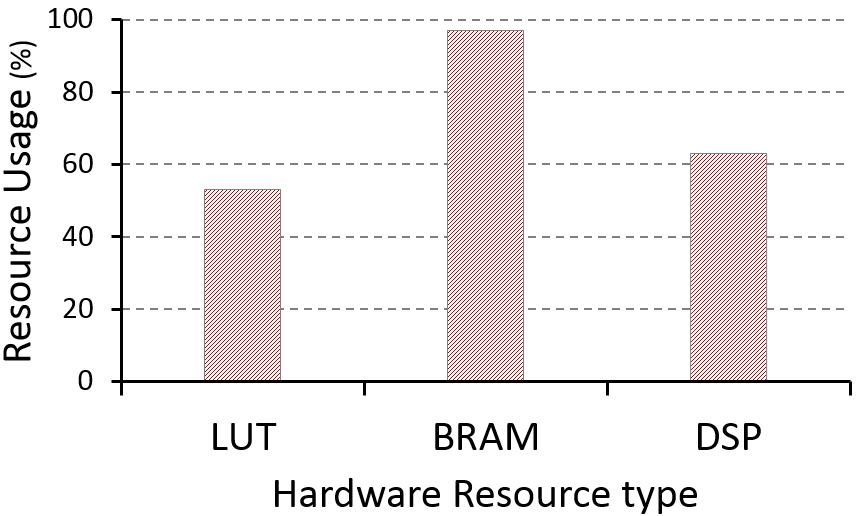}
        \caption{Hardware resource utilization of the Table~\ref{table:1} CNN implemented on FPGA.}\label{fig:res}
\end{figure}

\begin{table}
\centering
\caption{Software vs. TinyCNN accelerator.}\label{table:2}
\begin{tabular}{|p{1.3cm}||p{1.7cm}|p{1.7cm}|p{1.8cm}|}
\hline
 \centering version  & \centering Accuracy (\%) & \centering Runtime (ms) &\centering Data type  \tabularnewline 
\hline
\hline
 \centering SW &\centering 65.54 &\centering 42.54 &\centering 32-b floating \tabularnewline
\hline
 \centering HW-SM &\centering 62.28 &\centering 8.12 &\centering 16-b fixed \tabularnewline
\hline
 \centering HW-EM &\centering 62.28 &\centering 2.7 &\centering 16-b fixed \tabularnewline
\hline
\end{tabular}
\end{table}

Figure~\ref{fig:res} shows the resource usage of the CNN hardware implemented in \textit{PYNQ} board. As we expected, memory (BRAM) is the hardware resource limitation that we encounter while implementing a CNN on an embedded FPGA. In the software backend, the CNN architecture is designed in such a way that it utilized the FPGA BRAM to the highest.

\section{Conclusion}
In this paper, we proposed a framework that enables the designers to design a CNN accelerator for an embedded FPGA fast. The framework provides a software API that makes designers capable of exploring the CNN design space considering the hardware resource limitations. Then, the CNN hardware is generated and tuned to have a low accuracy loss by inter-layer precision adjustment. Our results show that we can reach up to $15.75\%$ speedup compared to the software implementation with 16-bit fixed-point data.

As future work, we are going to automate the software CNN design space exploration part in the software backend to make the CNN design even more easily. Since the only supported activation function is ReLU, we are going to add more activation units to the framework.

\bibliographystyle{ieeetr}
\bibliography{main1}

\end{document}